\pdfoutput=1

\documentclass[11pt]{article}

\usepackage[preprint]{acl}

\usepackage{makecell}
\usepackage{fvextra}
\usepackage{listings}
\usepackage{times}
\usepackage{latexsym}
\usepackage{hyperref}
\usepackage{lipsum}
\usepackage{times}
\usepackage{latexsym}
\usepackage{subfig}
\usepackage{graphicx}
\usepackage{subcaption}
\usepackage{amsmath}
\usepackage{multirow}
\usepackage{tabularx}
\usepackage{enumitem}
\usepackage{pbox}
\usepackage{float}
\usepackage{hyperref}
\usepackage{setspace}
\usepackage{diagbox}
\usepackage{float}
\usepackage{subcaption}
\usepackage{caption}
\usepackage{booktabs}
\usepackage{bbding}
\usepackage{mathtools}
\definecolor{grey}{rgb}{0.898,0.898,0.898}
\definecolor{ForestGreen}{rgb}{0.133,0.545,0.133}
\usepackage{amssymb}
\usepackage{pifont}
\usepackage{booktabs}      
\usepackage{array}         
\usepackage{amsmath}
\usepackage{booktabs}
\usepackage{multirow}
\usepackage{cleveref}
\usepackage{adjustbox}
\usepackage[T1]{fontenc}

\usepackage[utf8]{inputenc}

\usepackage{microtype}

\usepackage{inconsolata}

\usepackage{graphicx}

\title{Detecting Ambiguities to Guide Query Rewrite for Robust Conversations in Enterprise AI Assistants}

\author{Md Mehrab Tanjim$^1$, Xiang Chen$^1$, Victor S. Bursztyn$^1$, Uttaran Bhattacharya$^1$, Tung Mai$^1$,\\
\textbf{Vaishnavi Muppala$^2$, Akash Maharaj$^2$, Saayan Mitra$^1$, Eunyee Koh$^1$, Yunyao Li$^2$, Ken Russell$^2$}\\
$^1$Adobe Research, $^2$Adobe Inc.\\
  \texttt{\{tanjim, xiangche, soaresbu, ubhattac, tumai, mvaishna, maharaj,}\\ 
  \texttt{smitra, eunyee, yunyaol, kenrusse\}@adobe.com} \\}

\begin{document}
\maketitle
\begin{abstract}
Multi-turn conversations with an Enterprise AI Assistant can be challenging due to conversational dependencies in questions, leading to ambiguities and errors. To address this, we propose an NLU-NLG framework for ambiguity detection and resolution through reformulating query automatically and introduce a new task called ``Ambiguity-guided Query Rewrite.'' To detect ambiguities, we develop a taxonomy based on real user conversational logs and draw insights from it to design rules and extract features for a classifier which yields superior performance in detecting ambiguous queries, outperforming LLM-based baselines. Furthermore, coupling the query rewrite module with our ambiguity detecting classifier shows that this end-to-end framework can effectively mitigate ambiguities without risking unnecessary insertions of unwanted phrases for clear queries, leading to an improvement in the overall performance of the AI Assistant. Due to its significance, this has been deployed in the real world application, namely Adobe Experience Platform AI Assistant\footnote{\href{https://business.adobe.com/products/sensei/ai-assistant.html}{https://business.adobe.com/products/sensei/ai-assistant.html}}.

\end{abstract}

\section{Introduction}
Large Language Models (LLMs) have become increasingly popular and are now being integrated into many applications, such as document summarization \cite{kurisinkel2023llm, zakkas2024sumblogger},
information retrieval \cite{anand2023context, ma2023query}, conversational question answering (CQA) \cite{zhang2020dialogpt, thoppilan2022lamda, xu2023baize}, and so on.
Particularly in marketing and data analytics, LLMs can provide valuable information from documentation or general SQL queries to reveal data-related insights.
These digital experiences can be commonly referred to as AI Assistants \cite{maharaj2024evaluation}. 
These systems often involve multi-turn conversations that have conversational dependencies, including omissions, ambiguities, and coreferences \cite{anantha2021open, adlakha2022topiocqa, zhang2024clamber}, as illustrated in Figure 1\footnote{Please note throughout the paper we have used dummy names instead of the original ones for the customer specific dataset names and IDs for confidentiality.} (middle), where the second query, ``How many do I have?'' may not yield the correct answer without additional context from the first query ``What is a segment?" from the history.

\begin{figure}[tbp]
    \centering
    \includegraphics[width=\linewidth]{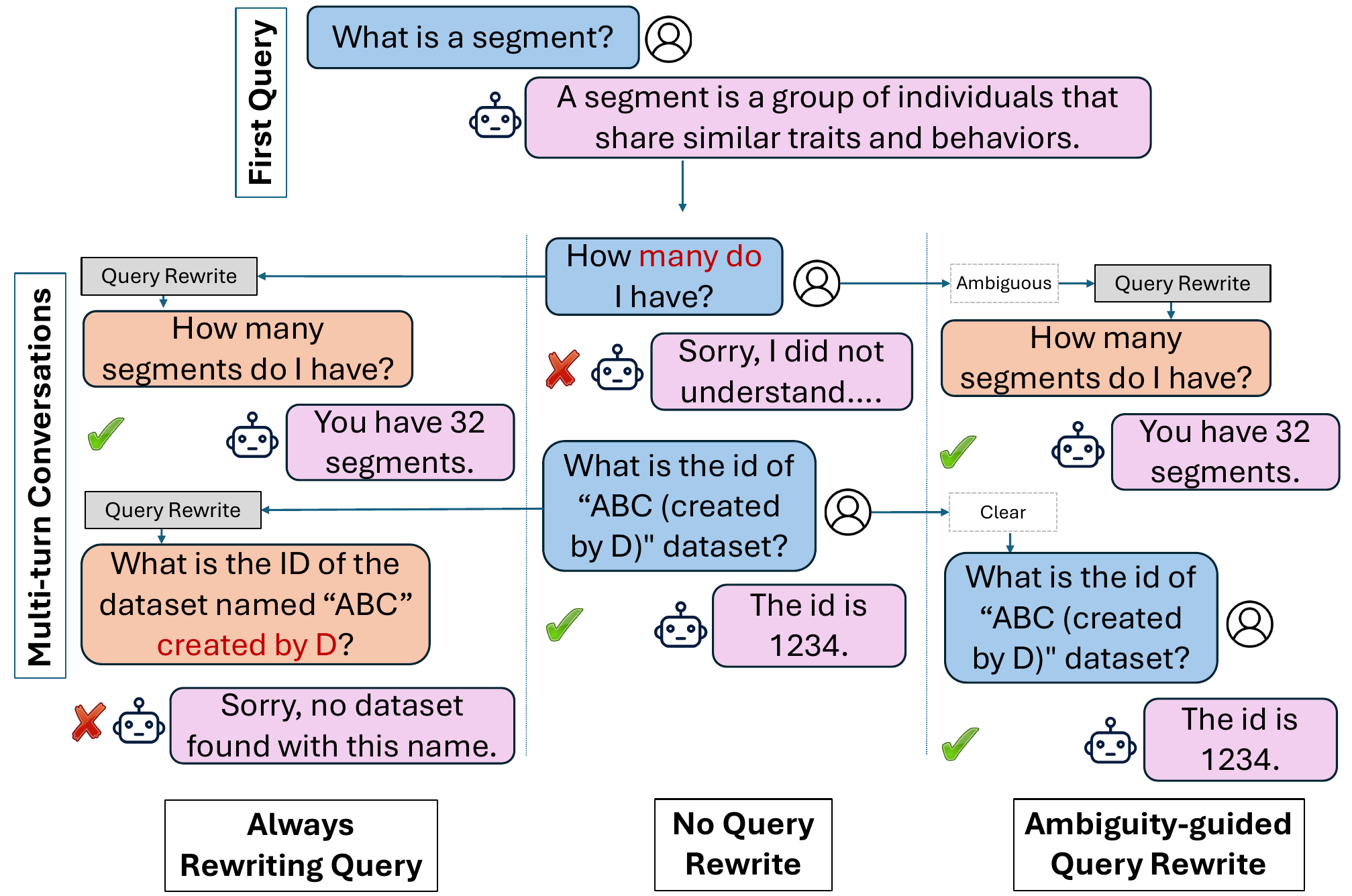} 
    \caption{Multi-turn conversations can have dependencies in prior chats, leading to ambiguities and errors (middle). While LLM-based rewriting can resolve some ambiguities, it may also introduce errors through unwanted rephrasing (left). Our proposed NLU-NLG framework, Ambiguity-guided Query Rewrite, rewrites only predicted unclear queries to prevent unnecessary rewrites leading to correct answers (right).}
    \label{fig:teaser}
\end{figure}

To disambiguate, an emerging approach is to prompt an LLM to rewrite the query based on previous chat history \cite{wang2023query2doc, jagerman2023query}.  
In the example in \Cref{fig:teaser} 
a prompted \texttt{GPT-3.5-Turbo}\footnote{\href{https://openai.com/}{https://openai.com/}}
for reformulating the query based on chat history can yield ``How many segments do I have?,'' which is a specific query that the AI assistant can answer correctly. 
This is known as Query Rewrite (QR) \cite{anand2023context, ma2023query}.
Prompting LLM for QR is a simple, yet effective solution to mitigate ambiguities in the query. However, rewriting all queries using LLM can lead to another problem. If we rewrite all queries by default, it can increase the risk of errors due to LLMs' tendency to hallucinate. This is shown in \Cref{fig:teaser} (left).
To mitigate this problem it is important to determine if a query rewrite is necessary or not. 
Especially if query is specific to begin with, it might be unnecessary to rewrite a query. This is illustrated in \Cref{fig:teaser} (right), where only `ambiguous' queries are rewritten and `clear' queries are bypassed leading to correct behavior for both of the queries.  By detecting unambiguous queries and skipping unnecessary rewrites, the chance of errors 
can be reduced ensuring an improvement in overall performance.

In this paper, we address this challenge
by first determining whether an incoming query is ambiguous or not using an ambiguity detecting classifier, which is our Natural Language Understanding (NLU) component and, if ambiguous, resolving it automatically by reformulating the query using a Natural Language Generation (NLG) component (e.g., an LLM). 
Specifically, our contributions in this paper are as follows:
\textbf{1)} We propose a novel NLU-NLG framework that addresses ambiguity detection and resolution through query rewriting and introduce a new task called ``Ambiguity-guided Query Rewrite'' (as shown in the right of \Cref{fig:teaser}) for robust multi-turn conversations in Enterprise AI Assistants. This task can serve as a standard approach for practitioners to build assistants using LLMs that can effectively handle ambiguous user queries. 
To the best of our knowledge, such a pipeline for disambiguation has not been explored before. 
\textbf{2)} We analyze the types of ambiguities that can arise in user conversational logs with AI Assistants and develop a taxonomy to understand the nature of these ambiguities. This taxonomy helps in creating specific detection systems tailored for each category of ambiguity.
\textbf{3) }
We use insights from our taxonomy to find useful features beyond text and devise rules for generating synthetic data points and improving detection. Our hybrid approach, which includes rule-based detection, data augmentation, and feature extraction, outperforms existing methods, including LLM-based baselines.
It is important to note that our proposed ambiguity-guided query rewrite pipeline is agnostic to any specific instance of the underlying QR model, making our solution 
ready to adopt in any industry setting. Because of its importance, our proposed pipeline has been integrated into Adobe Experience Platform (AEP) AI Assistant.


\section{Related Work}
\textbf{Query Rewrite.} 
Historically, Query rewrite or reformulation (QR) methods have included the addition of terms to the original query, known as query expansion \cite{lavrenko2017relevance}, or iteratively rephrasing the query using similar phrases \cite{zukerman2002lexical}, or synonymous terms \cite{jones2006generating}. 
Recently, the advent of large language models (LLMs) has spurred exploration into using these generative models to automatically resolve ambiguities during query processing.
For instance, recent studies have prompted LLMs to provide detailed information, such as expected documents or pseudo-answers \cite{wang2023query2doc, jagerman2023query}. These techniques are particularly effective when a golden dataset for a specific domain is unavailable, necessitating the use of off-the-shelf LLMs tailored for the specific use-case.
However, a
LLM-based QR can suffer from concept drift when using only queries as prompts \cite{anand2023context} and also has high inference costs during query processing. 
To address both, we introduce an ambiguous query understanding component to guide the rewrite process, ensuring that only unclear queries are rewritten, thereby limiting the undesired rewriting problem and saving cost at the same time. 

\begin{figure*}[htbp]
    \centering
    \includegraphics[width=\linewidth]{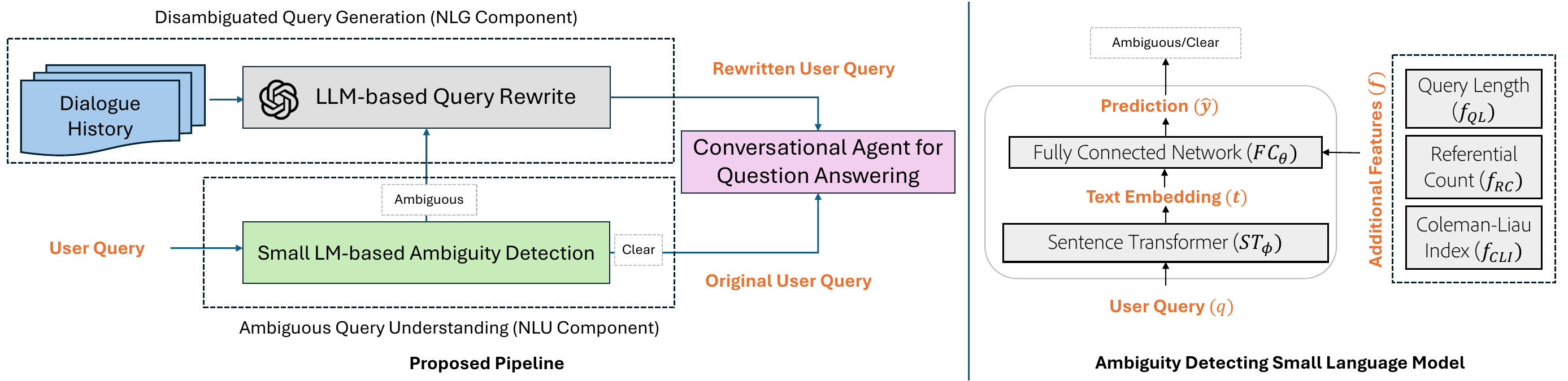} 
    \caption{Left: Proposed pipeline. Right: Architecture of our proposed ambiguity detection model.}
    \label{fig:pipline_and_architecture}
    \vspace{-0.5cm}
\end{figure*}

\textbf{Ambiguity Detection.} 
\label{sec:related_work}
Ambiguous queries are typically those that have multiple distinct meanings, insufficiently defined subtopics \cite{clarke2009overview}, syntactic ambiguities \cite{schlangen2004causes}, for which a system struggles to interpret accurately, resulting in inappropriate or unclear answers \cite{keyvan2022approach}. Detecting ambiguity in user queries, however, is a challenging task, as highlighted by several studies \cite{braslavski2017you, trienes2019identifying, guo2021abgcoqa}. Notably, \cite{trienes2019identifying} conducted the first comprehensive study on classifying questions as clear or unclear using logistic regression, especially in community platforms like Stack Exchange.
In CQA, \citet{guo2021abgcoqa} examined various ambiguities from a given text or story and also proposed methods for ambiguity detection and generating clarifying questions. 
In the era of large language models (LLMs), researchers like \cite{kuhn2022clam, zhang2024clamber} have shown that these models can be prompted to detect ambiguous questions. However, \citet{zhang2024clamber} pointed out the limitations of LLMs in accurately detecting them, which our experimental results also validate. 
In addition, all these research works focus on ambiguity detection for generating clarifying questions rather than query rewriting. In enterprise AI Assistants, however, avoiding clarifying questions is crucial for a smooth user experience, making automatic ambiguity resolution using QR
highly desirable. Furthermore, existing works share insights from open-domain datasets which might not always translate to industries as enterprise datasets are often close-domain. To that end, this paper investigates the effectiveness of combining ambiguity detection with query rewriting to resolve ambiguities automatically, sharing best practices for industry settings.

\section{Approach}

\subsection{Proposed Pipeline}

\Cref{fig:pipline_and_architecture} (left) illustrates our proposed ambiguity-guided query rewrite framework. In this system, the incoming query first passes through a Natural Language Understanding (NLU) component, i.e., an ambiguity detection classifier, to determine if it is ambiguous. If the query is clear, it is routed directly to the agent for conversational question answering. If ambiguous, it is rewritten by a Natural Language Generation (NLG) component for disambiguation. In our case, the NLG is a prompted LLM-based ``Query Rewrite'' (QR) module. 

\begin{table*}[ht]
\centering
\resizebox{\textwidth}{!}{
\begin{tabular}{p{1.5cm}p{6.5cm}p{3.5cm}p{7.5cm}}
\toprule
\textbf{Type} & \textbf{Definition} & \textbf{Example} & \textbf{Explanation} \\ 
\midrule
\multirow{2}{2cm}{Pragmatic\\(63.55\%)} & The meaning of a sentence depends on the context, reference, or scope. & \textit{``How many do I have?''} &  It is not clear what the user is referring to for the total count.\\ \hline
\multirow{2}{2cm}{Syntactic\\(31.90\%)} & The structure of a sentence is incomplete or allows for multiple interpretations. & \textit{``Business event''} &  It is not clear what the user is asking about ``business event.''\\ \hline
\multirow{2}{2cm}{Lexical\\(4.55\%)} & The meaning of the word/term is not clear or has multiple interpretations. & \textit{``Are we removing abc123 from XYZ?''} &  It is not clear what kinds of business objects (e.g., segment or dataset) \textit{abc123}/\textit{XYZ} are referring to.\\ 
\bottomrule
\end{tabular}}
\caption{The proposed taxonomy of ambiguous queries along with examples and explanations.}
\label{tab:taxonomy}
\vspace{-0.4cm}
\end{table*}

\subsection{Taxonomy of Ambiguous Queries}
To design a classifier that can detect unclear queries requiring rewriting, we need to first analyze the types of ambiguities that arise in real-world CQA systems. Categorizing these ambiguities can help us understand their nature and frequency, allowing us to prioritize the most common types and identify key signals for better feature design. Additionally, this analysis provides insights into generating synthetic data points to enhance model training. To that end, we analyzed about $3k$ user logs, with $400$ annotated as ambiguous, and developed a simple, yet effective taxonomy of three major ambiguity types: Pragmatic, Syntactical and Lexical, which are presented in \Cref{tab:taxonomy}. It is worth noting that while some taxonomies exist in the literature, such as \citet{zhang2024clamber}, these are often from open domains
and not derived from real-world deployed systems. Therefore, they may not provide relevant insights for an industry CQA system. For instance, they overlook syntactical and pragmatic ambiguities, which are predominant in our system. 
Our findings indicate that the most frequent ambiguities are pragmatic in nature, often referring to previous chats or responses. This insight allows us to create rules for synthetically generating additional data points for ambiguous queries, 
to circumvent the problem of having a low number of ambiguous queries (more on this later). We also notice that most of the lexical ambiguities follow a set of specific regular expression patterns for lexicons as well as a simple rule which makes the query ambiguous. We will touch on this next in our ambiguity detection discussion.

\subsection{Detecting Ambiguities}
One straightforward approach to detect ambiguities is to prompt-engineer another LLM to detect ambiguities. In our experiments, however, we have discovered that using an LLM to automatically detect ambiguities tended to label clear queries as ``ambiguous,'' lowering the Precision and limiting its applicability to resolve the aforementioned problem. To address this issue, we first draw some insights from our proposed taxonomy in \Cref{tab:taxonomy} and discover patterns.
Our developed taxonomy in \Cref{tab:taxonomy} enables us to devise two different approaches to handle the top-level types of ambiguities. Specifically, we discover that a small language model is capable of understanding whether a query has syntactical or pragmatic ambiguities, while rule-based detection of lexical ambiguities work quite well as from language modeling perspective there is nothing wrong with lexical ambiguous queries.

\subsubsection{Pragmatic and Syntactical Ambiguities}
\begin{figure}[tbp]
    \centering
    \includegraphics[width=\columnwidth]{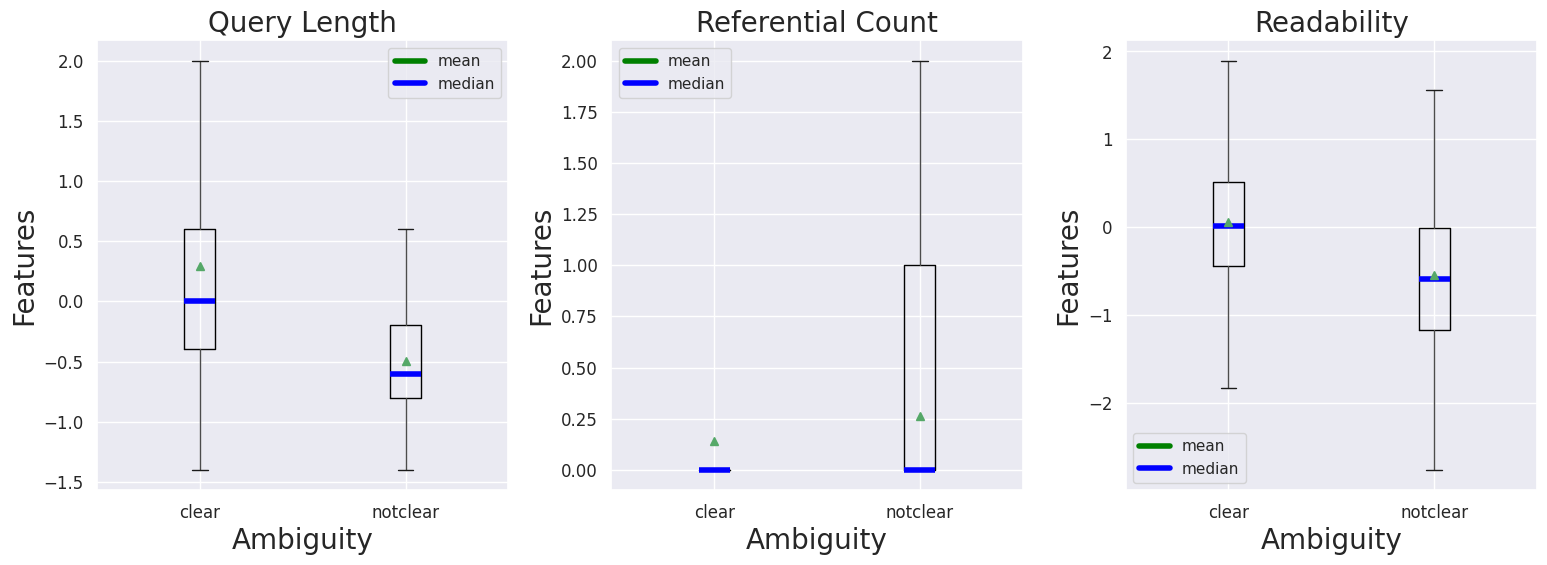} 
    \caption{The box plots of the features from our training set show a clear difference in distribution between clear and ambiguous queries\protect\footnotemark.}
    \label{fig:statistics_of_features}
    \vspace{-.3cm}
\end{figure}

To detect these types of ambiguous queries, our analysis of conversational logs led to an important insight that pragmatic ambiguities often arise due to usage of referential words
while syntactic ambiguities stem from inherent faults in the sentence structure. 
To capture both, we explore the following features: 1) Query Length ($f_{QL}$): total number of words in a query,  2) Referential Count ($f_{RC}$): total count of words from this list: [`this', `that', `those', `it', `its', `some', `others', `another', `other', `them', `above', `previous'], 3) Readability ($f_{CLI}$): Coleman-Liau Index (CLI) \cite{coleman1975computer} is a readability test designed to gauge the structure of a text, calculated by the following: $\text{CLI} = 5.89L/W - 30S/W - 15.8 $, where $L, S, W$  are the number of letters, words, and sentences respectively. \Cref{fig:statistics_of_features} shows the statistics which clearly show a pattern between clear and ambiguous queries. 

In addition to these features, the textual feature of the query itself plays a vital role.
For example, even readers who are not familiar with our system can distinguish between ``What is a segment?'' (clear) and ``What is it?'' (pragmatic ambiguity due to the usage of ``it'') or ``segment?'' (syntactic ambiguity due to grammatical structure). Therefore, we hypothesize that a language model would be better equipped to discern the difference, and our experiments have validated this hypothesis. 

\footnotetext{To account for outliers, we use robust scaling from \texttt{scikit-learn} \cite{scikit-learn} which removes the median and scales the data according to the quantile range.}

To incorporate both of these modalities of features, we first use a pretrained Sentence-Transformer \cite{reimers-2019-sentence-bert}, $ST_{\phi}$, where $\phi$ denotes the pretrained parameters as the underlying language model to get the text embedding, $\textbf{t}=ST_\phi(q)$, where $q$ is the query text. Then we use fully connected layers, $FC_{\theta}$ 
with $\theta$ parameters, 
which take both text embedding, $\textbf{t}$ and the robust normalized numerical feature vector, $\textbf{f}=[f_{QL}, f_{RC}, f_{CLI}]$, to get a final prediction, $\hat{y}=FC_{\theta}(\textbf{t}, \textbf{f})$. We train both $ST_\phi$ and $FC_{\theta}$ via backpropagation using cross entropy loss. Finally, since there is a class-imbalance, we use a weighted sampling during training. \Cref{fig:pipline_and_architecture} (right) shows the architecture of our ambiguity classifier. For more details, please see \Cref{sec:config}.


\subsubsection{Lexical Ambiguities}
To detect lexical ambiguities,
we observed that simple cases of data-related entities can easily be detected by a set of regular expressions, which led us to design simple, yet effective rules. Specifically,  
\begin{itemize}[noitemsep]
    \item Mask data-related entities by: 1) removing any weblinks, 2) filtering any ordinal numbers (like 1st, 2nd, etc.) and hyphen-separated words if they are commonly used in English (like pre-requisite), 3) matching words/phrases within a single or double quotation mark or with digits, periods, colon, underscore and dash, and finally 4) masking the matched words/phrases with ``ENTITY.''
    \item If the prediction from the model is ``clear'' but the entity types (which are pre-defined words from our business objects) are missing after masking, then label it as ``ambiguous.''
\end{itemize}
We show this rule in action with an example in \Cref{sec:lexicon}. A key takeaway from this exercise is that these rules can be modified and changed as per the requirements of the specific industry setting and based on customer data.

\subsection{Query Rewrite}
For the QR, we use
\texttt{GPT-3.5-Turbo} (version 1106)
However, it is important to note that our framework is not dependent on any specific LLM. Any prompted LLM 
can be used. Therefore, for generalizability of our solution, we also experiment with an open-source LLM, namely \texttt{Llama-3.1-70B}\footnote{\href{https://huggingface.co/meta-llama/Meta-Llama-3.1-70B}{https://huggingface.co/meta-llama/Meta-Llama-3.1-70B}}.
 There can be multiple approaches to designing a QR module. One simple prompt for QR could be as follows: \textcolor{ForestGreen}{``Rewrite the current query based on the previous chat history to remove any ambiguities.''}  While the specifics of our internal approach could not be shared due to legal constraints, we can offer valuable insights based on our model and prompt experimentation. In our prompt, we have used surrounding context, including the chat history of the past five interactions and relevant passage snippets. Then we have included 
instructions to transform the current user query into a fully specified query based on the context (e.g, chat history). We also have included instruction to
specifically address and clarify any ambiguous pronouns for co-reference resolution, correct typographical errors, and accurately preserve user-provided entity values. By following these guidelines, we enhance the accuracy and relevance of rewritten query in our AI Assistant.

\section{Experimental Setup}

\textbf{Data.} We collect user queries from the conversational logs with our AI Assistant and ask our domain experts to annotate these queries either `clear' or `ambiguous.'
Based on 
the conversational logs, our training set originally contained $3402$ queries, with $414$ being ambiguous,
and the test set had $489$ queries, of which only $84$ were ambiguous.
To address minority class imbalance in ambiguous queries for training,
we augment our training dataset by synthetically generating $1372$ queries based on insights from our taxonomy and rules from our lexical ambiguity detection. These rules include omitting proper nouns, randomly inserting referential words, creating vague statements, and masking queries by removing entity types. More details on this rule-based synthesis can be found in \Cref{sec:aug}.
For stress-testing our ambiguity detecting classifier for real-world deployment,  
we similarly augment our test set, but instead of using rules, human annotators were employed to come up with more ambiguous questions. The final test dataset contained a balanced set of $1036$ queries, with $403$ being unclear.
Finally, for comparing the query rewrite framework, we collect $366$ multi-turn conversations with golden rewrites, i.e., rewritten queries by human experts.
\begin{table}[tbp]  
    \centering  
    \resizebox{\columnwidth}{!}{
        \begin{tabular}{lcc}  
        \toprule
        \textbf{Model} & \textbf{F1} & \textbf{Accuracy}  \\
        \midrule
        SimQ\textsuperscript{$\star$} & 32.4 & 67.08 \\ 
        Abg-CoQA\textsuperscript{$\star\star$} & 73.44 & 83.25 \\
        Few-shot \texttt{GPT-3.5-Turbo}  & 71.58 & 72.68 \\
        Few-shot \texttt{Llama-3.1-70B}  & \underline{78.02} & \underline{79.14}\\
        \midrule
        \midrule
        Ours  & \textbf{90.19}$^\dagger$& \textbf{92.16}$^\dagger$ \\
        \bottomrule 
        \multicolumn{3}{l}{$\star$ based on \citet{trienes2019identifying}}\\
        \multicolumn{3}{l}{$\star\star$ based on \citet{guo2021abgcoqa}} \\
        \multicolumn{3}{l}{$\dagger$ results are statistically significant ($p<0.001$)} 
        \end{tabular}  
    }
\vspace{-0.2cm}
\caption{Comparison of performance across various models for detecting ambiguous queries.
}
\label{tab:comparison_ambiguity}
\vspace{-0.4cm}
\end{table}

\textbf{Baselines.}
For ambiguity detection, first baseline is a logistic regression model from \citet{trienes2019identifying} called ``SimgQ,'' which uses hand-selected features (such as query length).
The other baseline is referred to as ``Abg-CoQA'' from \citet{guo2021abgcoqa}, which treats ambiguity detection as a question answering task (mentioned in \Cref{sec:related_work}).
Additionally, inspired by \citet{zhang2024clamber}, we compare with LLM-based alternatives, where we prompt \texttt{GPT-3.5-Turbo} and \texttt{Llama-3.1-70B} with few-shot examples to detect ambiguities. 
More details on model configurations including ours and full prompt appear in \Cref{sec:config}.
To compare our proposed framework ``Ambiguity-guided Query Rewrite,'' we used two baselines: ``No Query Rewrites,'' where we do not have any QR, and ``Always Rewriting Query,'' where we always rewrite the query irrespective of its clarity.

\begin{table*}[t]
\centering
\resizebox{\textwidth}{!}{
\begin{tabular}{p{2.5cm}p{8.5cm}p{10cm}}
\toprule
\textbf{Original Query} & \textbf{Always Rewriting Query} & \textbf{Ambiguity-guided Query Rewrite} \\ 
\midrule
\textcolor{orange}{What is the id of ``ABC Dataset (created on)''?} 
& \texttt{\underline{GPT-3.5-Turbo:}}What is the id of the \textcolor{red}{``ABC Dataset''}?, \texttt{\underline{Llama-3.1-70B:}}What is the id of the \textcolor{red}{``ABC Dataset''?}; \underline{Response: }\textcolor{red}{Sorry there is no such dataset.} 
\vspace{0.2cm}
&  \underline{Predicted Ambiguity:} \textcolor{blue}{clear}; \textcolor{orange}{What is the id of ``ABC Dataset (created on)''?}; \underline{Response:} \textcolor{ForestGreen}{The id is 1234.} 
\\
\textcolor{orange}{What are its attributes?} 
& \texttt{\underline{GPT-3.5-Turbo:}} What attributes does the dataset \textcolor{red}{with the id} ``ABC Dataset (created on)'' have?, \texttt{\underline{Llama-3.1-70B:}} \textcolor{red}{What attributes would a dataset have?} 
&  \underline{Predicted Ambiguity:} \textcolor{blue}{ambiguous}; \texttt{\underline{GPT-3.5-Turbo:}} What are the attributes of \textcolor{ForestGreen}{dataset 1234?}, \texttt{\underline{Llama-3.1-70B:}} What are the attributes of dataset \textcolor{ForestGreen}{``ABC Dataset (created on)'' with id 1234?}\\
\bottomrule
\end{tabular}}
\vspace{-0.2cm}
\caption{Illustrative examples on how our ambiguity-guided query rewrite leads to the correct behavior.}
\label{tab:qualitative}
\vspace{-0.4cm}
\end{table*}

\begin{table}[tbp]  
    \centering  
    \resizebox{\columnwidth}{!}{
        \begin{tabular}{ccccc|c|c}  
        \toprule
        \multicolumn{5}{c|}{\textbf{\texttt{all-mpnet-base-v2}}} & \makecell{\textbf{\texttt{all-distil}}\\\textbf{\texttt{roberta-v1}}} &  \makecell{\textbf{\texttt{all-MiniLM}}\\\textbf{\texttt{-L12-v2}}} \\
         \textbf{Base} & \textbf{Base+R} & \makecell{\textbf{Base+R}\\\textbf{+AD}} & \makecell{\textbf{Base+R}\\\textbf{+AD+AF}} & \makecell{\textbf{All}\textsuperscript{$\star$}} & \makecell{\textbf{All}\textsuperscript{$\star$}} & \makecell{\textbf{All}\textsuperscript{$\star$}}\\
        \midrule
         88.29 & 89.92  & 91.91 & 92.04 & \textbf{93.37} & 92.33 & 92.36 \\ 
        \bottomrule 
        \multicolumn{5}{l}{$\star$ All=Base+R+AD+AF+WS}\\
        \end{tabular}  
    }
\caption{Recall for ``ambiguous'' queries shows the contribution of using rules (R), augmented data (AD), additional feature (AF) and weighted sampling (WS).}
\label{tab:ablation_study}
\vspace{-0.5cm}
\end{table}


\textbf{Metrics.} For ambiguity detection, we use F1, and accuracy. For experiments related to the framework, we use cosine similarity\footnote{using \texttt{text-embedding-3-large} from OpenAI}, BERTScore \cite{zhang2019bertscore}, and BLEU score (average of 1 \& 2-gram) to measure similarities with ground truth rewritten queries. 

\section{Experiments and Results}




We compare the performance of our ambiguity detection model with several baselines and show the results in \Cref{tab:comparison_ambiguity} (average numbers are reported over 10 runs).
Our findings indicate that SimQ, which uses hand-picked features, resulted in poorest performance. This is expected, as textual features which play a vital role to understand the query, were not used by \citet{trienes2019identifying}. On the other hand, Abg-CoQA saw a slight improvement in using textual features, but its performance in both F1 and accuracy is still lower overall. This result aligns with their original finding \cite{guo2021abgcoqa} that ambiguity detection as a question answering task has its limitations. 
LLM-based baselines did better than previous two but still the overall performance is not satisfactory, especially for deployment. This relatively lower
performance of LLM-based ambiguity detection in our case
also aligns with the study in
\citet{zhang2024clamber}. Our proposed ambiguity detection model, on the other hand, outperforms all of the baselines under both F1 and Accuracy. \Cref{tab:ablation_study} demonstrates that each introduced component, such as rules for masking and detecting lexical ambiguities, additional features, and weighted sampling for class imbalance, contributed to performance gains. In addition, \texttt{all-mpnet-base-v2} provided the best performance compared to other pre-trained sentence-transformers, consistent with its leaderboard ranking (\href{https://www.sbert.net/docs/sentence_transformer/pretrained_models.html}{sbert.net}).
Since ambiguity-detecting has a higher performance, we can expect the ambiguity-guided rewrite to reduce hallucinations or unwanted rewrites, which we find to be the case in our experiments; \Cref{tab:framework_comparison} shows our proposed framework indeed outperforms all alternatives across all metrics.
This is because the ground truth rewritten query is close to the original query if it is clear and thus by skipping QR if detected clear, the output semantically matches more with ground truths.
Since similarity metrics alone cannot capture the cascading effect of wrong rewrites, we also demonstrate an illustrative example inspired by a real conversation from one of our users in \Cref{tab:qualitative}. 
In these examples, the first query asks for the ID of the dataset named ``ABC Dataset (created on),'' but QR without guidance incorrectly removed ``(created on)'' from the name, leading to an incorrect response. Normally such small 
rephrasing would not be an issue, however, in the context of generating the correct SQL query, this can lead to a significant error, like dataset not found. 
The previous mistake of the incorrect rewriting of the dataset name then leads to another incorrect rewrite in the second turn, where the LLM mistakes the name as ID (\texttt{GPT-3.5-Turbo}) or leads to a wrong question all together (\texttt{Llama-3.1-70B}). Our proposed framework avoids such mistakes through guided rewriting as highlighted in green, resulting in correct behavior in all cases.
Not surprisingly, after productionizing our proposed pipeline, the error related to downstream tasks (such as routing to proper agents, generating SQL queries, etc.) in our application reduced from \textbf{18\% to 8\%}. As an additional outcome, since the latency of our ambiguity classifier is merely \textbf{0.01 seconds}, our ambiguity-guided rewrite approach lowers both costs and latency associated with LLM calls by bypassing QR for clear queries. 
Overall, our quantitative and qualitative results highlight the importance of 
providing guidance for query rewrites for improve the performance of the overall system.

\begin{table}[t]  
    \centering  
    \resizebox{\columnwidth}{!}{
        \begin{tabular}{lccc|ccc}  
        \toprule
         \multirow{5}{*}{\makecell{\textbf{Disambiguation} \\\textbf{Frameworks}}} & \multicolumn{3}{c|}{\textbf{\texttt{GPT-3.5-Turbo}}} & \multicolumn{3}{c}{\textbf{\texttt{Llama-3.1-70B}}} \\
         & \rotatebox{90}{\textbf{BLEU Score}} & \rotatebox{90}{\textbf{Cosine Sim.}} & \rotatebox{90}{\textbf{BERTScore}} & \rotatebox{90}{\textbf{BLEU Score}} & \rotatebox{90}{\textbf{Cosine Sim.}} & \rotatebox{90}{\textbf{BERTScore}}  \\
        \midrule
        No Query Rewrite & 0.47 & 0.6863 & 0.8187  & 0.47 & 0.6863 & 0.8187 \\
        Always Rewriting Query & 0.4755 & 0.8274 & 0.8529 & 0.4301 & 0.8089 & 0.8441 \\
        Ambiguity-guided Rewrite  & \textbf{0.5417} & \textbf{0.847} & \textbf{0.8680 } & \textbf{0.5314}  & \textbf{0.8410} & \textbf{0.8653} \\
        \bottomrule 
        \end{tabular}  
    }
\vspace{-0.3cm}
\caption{Quantitative comparison with golden rewrites show the effectiveness of our proposed framework. For fair comparison, we use the same prompt in both LLMs.}
\label{tab:framework_comparison}
\vspace{-0.5cm}
\end{table}

\vspace{-0.1cm}
\section{Conclusion}
\vspace{-0.1cm}
In this paper, we have explored the challenges of ambiguity in conversations with LLM-based agents
in the enterprise settings. To mitigate this challenge and ensure the correct behavior of such intelligent systems, we have proposed 
an NLU-NLG framework for disambiguation through ambiguity-guided query rewriting. Our approach, which combines rules and models with a mix of hand-picked and textual features, effectively detects ambiguous queries. 
Furthermore, our findings demonstrate that query rewriting guided by detected ambiguity results in superior performance for disambiguation and robust conversations with correct behavior, ultimately leading to its deployment into our product. Overall, we hope our insights will provide valuable guidance for researchers and practitioners working on conversational systems and AI Assistants. 


\appendix

\section{Model Configurations and Prompt }
\label{sec:config}
\textbf{Baselines.} For SimQ, we use the code provided by \citet{trienes2019identifying}. Their model takes the title and body of a Stack Exchange post as well as corresponding clarifying questions. Since we only have query texts, we provide the query text as input to the title and leave other fields empty. For Abg-CoQA, we use the setup mentioned in \citet{guo2021abgcoqa} and pretrain BERT-base model on CoQA dataset \cite{reddy2019coqa} and their proposed Abg-CoQA dataset and then fine tune on our dataset. They frame ambiguity detection as question answering task which needs a question, chat history, and input passage, where they insert ``ambiguous'' in the end for the ambiguous query. Following their setup, we also give the query and chat history as inputs, but since we do not have any passage, we use `unknown. clear. ambiguous' as input passage so that it would extract `clear' for clear queries, `ambiguous' for ambiguous queries, and `unknown' if the query does not have label (which happens for queries in the chat history).

\textbf{Our model.} For our model, we use three different backbones from the pretrained Sentence Transformer models which have the highest sentence embeddings performance in the leaderboard on \href{https://www.sbert.net/docs/sentence_transformer/pretrained_models.html}{https://www.sbert.net/docs/ sentence\_transformer/pretrained\_models.html}: \texttt{all-mpnet-base-v2}, \texttt{all-distilroberta-v1} and \texttt{all-MiniLM-L12-v2}. From them, the first two outputs $768$ dimensional vector for the text while the last one outputs $384$ dimensional vector. As such the input dimension for our fully connected layers $FC_\theta$ also change accordingly based on the backbone. $FC_\theta$ has two layers, the input-output size of first layer is ($768+3$, $384$) for \texttt{all-mpnet-base-v2} and \texttt{all-distilroberta-v1} or ($384+3$, $384$) for \texttt{all-MiniLM-L12-v2} (the additional $3$ is due to the $3$-dimensional feature vector for query length, referential count, and readability), followed by a \texttt{Tanh} activation, a dropout layer and a final fully connected layer  of ($384$, $2$) for the prediction. For training we use a learning rate of $2e-05$, batch size of $4$, and $3$ epochs with Adam optimizer. We also use a weighted sampling to have a balanced number of `clear' and `ambiguous' queries in each batch ($1:1$ ratio). We calculate validation loss every $50$ steps and save the best model which gives the highest average of both Recall and F1 for the ``ambiguous'' class.

\textbf{Prompt template for LLM-based detection.} We use the following prompt to detect ambiguity in query. Please note any specific word/phrase belonging to our internal data in few-shot examples has been censored with `***' for confidentiality. 

\begin{Verbatim}[breaklines=true]
prompt = f'''You are an expert linguist. You are provided with a question that a customer is asking to a conversational assistant, your task is to evaluate the clarity of question. Note that: 

* If the current question is complete and customer's intent is clear, output "RESPONSE: CLEAR" on a separate line. 

* If the current question is smart-talk or chit-chat, output "RESPONSE: CLEAR" on a separate line.

* If the current question is ambiguous or need clarification from the customer, output "RESPONSE: VAGUE" on a separate line. 

* If the current question contains coreference and becomes clear after coreference resolution, output "RESPONSE: VAGUE" on a separate line. 

* if the current question contains missing pronouns or conditions based on the conversation history, output "RESPONSE: VAGUE" on a separate line. 

* Do not include any explanation. 

Examples: 

QUESTION: How can I connect *** with ***? 
RESPONSE: CLEAR 

QUESTION: What is the *** from *** to populating in ***? 
RESPONSE: CLEAR 

QUESTION: When should I use the product? 
RESPONSE: VAGUE 

QUESTION: what is the data retention policy in ***? 
RESPONSE: CLEAR 

QUESTION: What is the license in this case? 
RESPONSE: VAGUE 

QUESTION: what's ***? 
RESPONSE: CLEAR 

QUESTION: what does this page do? 
RESPONSE: VAGUE 

CURRENT QUESTION:{curr_query}'''
\end{Verbatim}

\section{Lexical Ambiguity Detection with an Example}
\label{sec:lexicon}

When we match any term that fit into our patterns, we mask them by ``ENTITY.'' An example of this conversion is: \textit{``What is the total size of 124abcde?''}, which will be converted to \textit{``What is the total size of ENTITY?''} This further enable us to devise a simple rule to detect lexical ambiguities. The rule is as mentioned in the paper: \textcolor{orange}{If a masked query has an ENTITY in it but does not have any words for the ENTITY types, then it has a lexical ambiguity.} 

For looking up entity types, we have a pre-defined list of words, which is internal to our company, but for example's sake, we can use the entity types mentioned in the paper such as `segment', `schema' or `dataset.'  

We have found this rule to be quite effective for detecting simple cases of lexical ambiguities. For example, in the above case, the query will be detected as `ambiguous' as it is not clear what the entity is referring to (it could be a dataset or segment or something else). 

  

\section{Data Augmentation}
\label{sec:aug}
We synthetically generate total $1372$ queries that needs a rewrite by following these rules: 1) omit any proper nouns, 2) insert referential pronouns randomly and 3) creating vague statements using phrases like `there is no such', `there is not any' etc. Specifically: 

1) Omitting Details: In this rule, we first match any sentences with the regex match `the (\textbackslash w+) of', which will find the patter ``the \{\} of''. Then we remove the words after ``of'' (inclusive), to make the sentence vague. For example, ``What is the name of my largest dataset?'' to ``What is the name?'' In this fashion, we generate $80$ unclear queries. 

2) Adding Referential Pronouns: In this rule, we first use the $80$ generated unclear queries above where it has the word ``the'' followed by a noun. Then we find the occurrences of ``the'' and replace it randomly with any of the referential words from this list:  \textcolor{orange}{[`this,' `that, `those,' `it,' `its,' `some,' `others,' `another,' `other,' `above', `previous']}. For example, applying this rule to ``What is the name?'' can generate ``What is this name?.'' We repeat the process $5$ times yielding $270$ unclear queries. Furthermore, we take all the clear queries with sentence length less than or equal to $7$ and then apply the same rule $5$ times giving us $353$ more unclear queries. Total unclear queries from this rule is: $623$. 

3) Vague Statements: For this rule, we first filter all the queries which are non-questions (detected by the absence of `?' in the query). Then, using parts-of-speech tagger, we find all the sentences that start with a verb, followed by a pronoun. Then we replace both verb and the pronoun randomly with any of the following phrases: \textcolor{orange}{[`there is', `there are', `there is no such', `there is no', `there are no such', `there are no', `there is not any,' `it is', `it is not', `this is not', `this is', `that is', `that is not']}. For example, applying this rule will find statements like ``Tell me about `ABC' dataset'' and generate vague statement like ``There is no such `ABC' dataset''. Repeating this process $5$ times for each sentence yields $669$ vague statements.

\end{document}